\documentclass{ecai} 
\usepackage{latexsym}
\usepackage{amssymb}
\usepackage{amsmath}
\usepackage{amsthm}
\usepackage{booktabs}
\usepackage{enumitem}
\usepackage{graphicx}
\usepackage{color}

\usepackage{subcaption}
\usepackage{multirow}
\usepackage{xspace}
\usepackage{booktabs}
\usepackage{amsmath}
\usepackage{xcolor}
\usepackage{bm}
\DeclareMathOperator*{\argmax}{argmax}

\newcommand\matteo[1]{\textcolor{magenta}{Matteo: #1}}

\newcommand{\system}{MToMnet\xspace}

\newcommand{\BibTeX}{B\kern-.05em{\sc i\kern-.025em b}\kern-.08em\TeX}

\begin{document}

%%%%%%%%%%%%%%%%%%%%%%%%%%%%%%%%%%%%%%%%%%%%%%%%%%%%%%%%%%%%%%%%%%%%%%%%

\begin{frontmatter}

%%% Use this command to specify your submission number.
%%% In doubleblind mode, it will be printed on the first page.

\paperid{123} 

%%% Use this command to specify the title of your paper.

\title{Explicit Modelling of Theory of Mind for Belief Prediction in Nonverbal Social Interactions}

%%% Use this combinations of commands to specify all authors of your 
%%% paper. Use \fnms{} and \snm{} to indicate everyone's first names 
%%% and surname. This will help the publisher with indexing the 
%%% proceedings. Please use a reasonable approximation in case your 
%%% name does not neatly split into "first names" and "surname".
%%% Specifying your ORCID digital identifier is optional. 
%%% Use the \thanks{} command to indicate one or more corresponding 
%%% authors and their email address(es). If so desired, you can specify
%%% author contributions using the \footnote{} command.

\author[A]{\fnms{Matteo}~\snm{Bortoletto}
%\orcid{....-....-....-....}
\thanks{Corresponding Author. Email: matteo.bortoletto@vis.uni-stuttgart.de.}}
\author[A]{\fnms{Constantin}~\snm{Ruhdorfer}
%\orcid{....-....-....-....}
}
\author[A]{\fnms{Lei}~\snm{Shi}
%\orcid{....-....-....-....}
} 
\author[A]{\fnms{Andreas}~\snm{Bulling}
%\orcid{....-....-....-....}
}

\address[A]{University of Stuttgart, Germany}

%%% Use this environment to include an abstract of your paper.

\begin{abstract}
    We propose \textit{\system}~-- a Theory of Mind (ToM) neural network for predicting beliefs and their dynamics during human social interactions from multimodal input.
ToM is key for effective nonverbal human communication and collaboration, yet existing methods for belief modelling have not included explicit ToM modelling or have typically been limited to one or two modalities.
\system encodes contextual cues (scene videos and object locations) and integrates them with person-specific cues (human gaze and body language) in a separate \textit{MindNet} for each person.
Inspired by prior research on social cognition and computational ToM, we propose three different \system variants: two involving the fusion of latent representations and one involving the re-ranking of classification scores.
We evaluate our approach on two challenging real-world datasets, one focusing on belief prediction while the other examining belief dynamics prediction. 
Our results demonstrate that \system surpasses existing methods by a large margin while at the same time requiring a significantly smaller number of parameters.
Taken together, our method opens up a highly promising direction for future work on artificial intelligent systems that can robustly predict human beliefs from their non-verbal behaviour and, as such, more effectively collaborate with humans.

\end{abstract}

\end{frontmatter}

\section{Introduction}

Social interaction and collaboration are essential human skills~\citep{floyd2011interpersonal}.
To engage in them effectively, humans have developed the ability to predict mental states and beliefs of others by observing their nonverbal behavioural cues, such as gaze or body language -- so-called Theory of Mind~\citep[ToM]{premack1978does}.
Humans are adept at integrating multiple modalities for this task, including contextual information. 
Given its importance in human-human interactions, computational ToM has recently emerged as a new frontier in developing intelligent computational agents that can understand and collaborate with humans~\citep{gurney2022robots}.
Despite a surge of papers on this new task, deep learning methods for predicting other agents' mental states have mainly been studied in constrained artificial environments~\citep{baker2017rational, rabinowitz2018machine, nguyen2020cognitive, gandhi2021baby, shu2021agent, sclar2022symmetric, nguyen2022learning, nguyen2023memory, bortoletto2024neural}. 
Moreover, existing methods typically rely on only one or a few modalities to predict beliefs, such as visual or linguistic cues~\citep{liu2023computational, takmaz2023speaking}. 
Effectively integrating a wider range of modalities for belief prediction remains an open research challenge. 
Moreover, existing approaches for belief prediction in real-world settings have not explicitly added a ToM mechanism.

In this work, we focus on predicting human beliefs from multimodal inputs and how these beliefs change dynamically in real-world scenarios involving naturalistic dyadic (human-human) interactions.
Belief prediction is particularly challenging, and ToM is particularly important when verbal communication is impossible.
In these situations, individuals must instead rely on nonverbal cues to convey their intentions and beliefs. 
Advancing from recent work~\citep{fan2021learning, duan2022boss}, we propose a multimodal ToM neural network (\textit{\system}) that leverages person-specific nonverbal communication cues (gaze, pose) and contextual cues (video frames, object bounding boxes) to predict beliefs and how they change over time.

\system encodes contextual cues using shared feature extractors and person-specific cues using two independent \textit{MindNet}s -- LSTM-based sub-networks that allow our model to encode individual traits.
Without a clear theoretical framework for integrating ToM into neural networks, we draw inspiration from computational ToM and social cognition research and study three different variants of \system that add explicit ToM modelling. 
The Decision-Based \system (DB-\system) adopts a decision-based strategy inspired by recent advancements in referential games~\citep{liu2023computational}.
Here, belief prediction for one individual is used to re-rank the predictions for the other. 
The other two approaches employ a model-based strategy, leveraging 
MindNets' latent representations.
The Implicit Communication \system (IC-\system) 
enables the communication between MindNets via late fusion of internal representations.
The Common Ground \system (CG-MToMnet) is inspired by research in social cognition, in particular by the idea that human communication involves a shared, inter-subjective \textit{common ground}~\citep{tomasello2010origins}.
We use \system latent representations to create such common ground based on this insight.

We evaluate these \system variants on two challenging multimodal real-world datasets that target complementary objectives.
The Benchmark for Human Belief Prediction in Object-context Scenarios \citep[BOSS]{duan2022boss} consists of videos of two people tasked to
collaborate only using nonverbal communication.
BOSS facilitates the evaluation of models' \textit{belief prediction} capabilities, i.e., the ability to correctly predict the belief of both people for each video frame.
In contrast, the Triadic Belief Dynamics dataset \citep[TBD]{fan2021learning} focuses on 
communication events that emerge during in-the-wild social interactions between two people. 
TBD enables the evaluation of models' ability to predict changes in the \textit{belief dynamics} of a person causally constructed by these events.

We report extensive experiments on both datasets, demonstrating that our approach significantly outperforms state-of-the-art methods while only using a fraction of the parameters. 
Our results emphasise the importance of explicit Theory of Mind (ToM) modelling for achieving these performance improvements.
Moreover, analyses of \system 's latent representations underline the effectiveness of encoding person-specific cues with independent MindNets.
Further post-hoc analyses on TBD show that \system can predict \textit{false beliefs dynamics} -- beliefs that do not align with reality -- more accurately than previous approaches.

Overall, this work makes four contributions:
\begin{itemize}
    \item We introduce a multimodal ToM neural network (\system) that combines nonverbal communication cues and visual inputs for belief and belief dynamics prediction.
    \item We propose three approaches to computationally modelling Theory of Mind, inspired by recent work on computational ToM and social cognition: decision-based, implicit communication, and common ground.
    \item We demonstrate that explicit ToM modelling allows us to achieve substantial performance gains for two tasks -- belief prediction and belief dynamics prediction -- at a significantly lower computational cost.
    \item We report analyses highlighting the effectiveness of modelling person-specific beliefs using independent MindNets and the efficacy of explicit ToM modelling for capturing false beliefs.
\end{itemize}

\begin{figure*}[t]
    \centering
    \includegraphics[width=\textwidth]{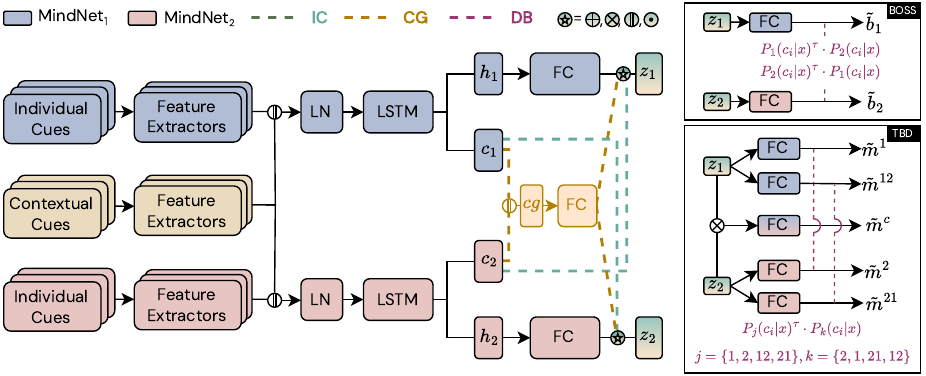}
    \caption{Our multimodal Theory of Mind neural network \textit{\system} consists of two separate \textit{MindNet}s -- one for each person -- to encode individual cues (e.g.\ human gaze and body language) and integrate them with contextual cues (e.g.\ scene videos and object locations). We propose three different \system variants:
    Decision-Based \system (DB-\system) combines class probabilities of the two MindNets to re-rank predictions. Implicit Communication \system (IC-\system) adds a communication mechanism between the two MindNets that exchange their internal LSTM cell state. Common Ground \system (CG-\system) forms a common ground representation by concatenating the two MindNets' cell states. We also study different aggregation operations: $\otimes$ = element-wise multiplication, $\oplus$ = element-wise sum, $\parallel$ = concatenation, and $\odot$ = cross-attention.
    }
    \vspace{0.1cm}
    \label{fig:arch}
\end{figure*}

\section{Related Work}

\subsection{Belief Prediction}

Predicting beliefs is a challenging task, even in constrained artificial settings, such as grid-world environments~\citep{rabinowitz2018machine, gandhi2021baby, sclar2022symmetric, nguyen2022learning, nguyen2023memory, bortoletto2024neural}, 3D worlds consisting of basic geometric shapes~\citep{shu2021agent} or virtual reality environments~\citep{puig2020watch}.
First, datasets that take a step towards mental state modelling in real-world settings have been proposed.
These datasets consist of videos of human social interactions that have been annotated with rich social cues, such as gaze or body pose.
The Benchmark for Human Belief Prediction in Object-context Scenarios (BOSS) focuses on \textit{belief prediction} in dyadic collaborative interactions, i.e.\, the task of predicting beliefs of two people collaborating with each other \citep{duan2022boss}.
Similarly, the Triadic Belief Dynamics dataset (abbreviated TBD here) focuses on the prediction of \textit{belief dynamics}, i.e.\ predicting if and how someone's belief changes during social nonverbal interactions \citep{fan2021learning}.
As such, both datasets complement each other in terms of the tasks they evaluate and the scenarios they cover, namely in-the-wild everyday activities (TBD) and collaborative scenarios (BOSS). 
In this work, we use both datasets to evaluate our method and show that employing a triadic structure and explicitly modelling ToM achieves better performance than existing methods for predicting both beliefs and belief dynamics.

\subsection{Machine Theory of Mind}

Theory of Mind (ToM) has been studied in cognitive science and psychology for decades, but our understanding of how humans develop this essential ability is still severely limited.
Mirroring efforts to understand ToM in humans, an increasing number of works in the computational sciences have investigated means to equip artificial intelligent (AI) systems with similar capabilities.
Previously proposed models that aim to implement a machine ToM have been based on partially observable Markov decision processes (POMDP)~\citep{doshi2010modeling, han2018learning}, Bayesian methods~\citep{baker2009action, lee2019bayesian, fan2021learning, netanyahu2021phase} and deep learning methods~\citep{rabinowitz2018machine, bara2021mindcraft, wang2022tomc, duan2022boss, liu2023computational, bortoletto2024neural, bortoletto24Limits, bortoletto2024benchmarking}.
Specifically for predicting beliefs of agents that engage in nonverbal communication, \citet{duan2022boss} and \citet{fan2021learning} follow different approaches.
\citet{duan2022boss} have used deep learning methods based on a ResNet~\citep{he2016deep} feature extractor for video frames and linear feature extractors for gaze, pose, bounding boxes and object-context relations. 
In contrast, \citet{fan2021learning} has used a triadic hierarchical energy-based model to track beliefs dynamics and compared it to neural network baselines that use only RGB frames, histogram of oriented gradients~\citep[HOG]{dalal2005histograms} or handcrafted features. 
Current deep learning approaches either handle input modalities shallowly or restrict themselves to a limited set of modalities.
Energy-based models rely on more upfront engineering work involving the use of handcrafted features.
Moreover, none of these approaches models ToM explicitly in their formulation.
In this work, we show how explicitly modelling ToM in the neural network architecture can lead to substantial improvements compared to previous approaches.

\section{Method}

Our multimodal Theory of Mind neural network (\system) combines nonverbal human communication cues (gaze and pose) with 
contextual cues (e.g.\ RGB video frames and object bounding boxes) to predict the beliefs of two observed human agents. 
In stark contrast to previous approaches~\citep{fan2021learning, duan2022boss}, our method leverages shared feature extractors and two \textit{MindNet}s that individually model each person's beliefs (see Figure~\ref{fig:arch}). 
This design choice is motivated by research in social cognition suggesting that triadic (human-human-context) joint attentional engagement is necessary for effective cooperation~\citep{tomasello2010origins}.
This triadic engagement
is reflected by our choice of two MindNets that encode nonverbal cues of each human and shared feature extractors that encode contextual cues available to both humans.

\subsection{Base MToMnet}
Our base \system consists of two separate MindNets and a set of shared feature extractors.
Each MindNet encodes individual cues from one person (e.g.\ human gaze and body language), combines them with contextual features (e.g.\ scene videos and object locations) coming from the shared features extractors, and adds temporal information.
Let $\bm{x}_C\in\mathcal{C}$ be a set of contextual cues and $\bm{x}_I\in\mathcal{I}$ a set of individual cues for a specific person in the scene. 
Contextual cues are encoded by shared features extractors and concatenated
\begin{equation}
    \bm{x}_{ctx} = \big\Vert_{C\in \mathcal{C}} \texttt{SharedFeatExtr}_C(\bm{x}_C)
\end{equation}
where $\parallel$ denotes concatenation. 
Similarly, individual cues for a particular person are encoded and concatenated:
\begin{equation}
    \bm{x}_{ind} = \big\Vert_{I\in \mathcal{I}} \texttt{MindNetFeatExtr}_I(\bm{x}_I)
\end{equation}
Individual and contextual features are subsequently concatenated and used as input
to a normalisation layer~\citep[LN]{ba2016layer}, followed by a bidirectional LSTM~\citep{graves2005framewise} to model temporal information. 
The final LSTM hidden state is passed on to one or more fully connected (FC) classification heads that yield a probability distribution over classes:
\begin{align}
    P(y_i|\bm{x}) \propto \exp(\texttt{FC}(\texttt{LSTM}(\texttt{LN}(\bm{x}))))
\end{align}
where $\{y_i\}_{i=0}^{Y}$ are dataset-specific classes and
$\bm{x} = \bm{x}_{ctx} \parallel \bm{x}_{ind}$.
The final belief predictions are obtained using the argmax of $P(y_i| \bm{x})$:
\begin{align}
    \Tilde{b} &= \argmax_i P(y_i|\bm{x}) 
\end{align}
In the following, we refer to the inputs, latent representations, and outputs of the two person-specific MindNets with the indices $\{1,2\}$.
We investigate three different variants of explicitly adding Theory of Mind to this base model architecture.

\subsection{\system Variants}
We study three different variants of \system that \textit{draw inspiration} from prior research on computational ToM and social cognition to add explicit ToM modelling: Decision-Based (DB-\system), Implicit Communication (IC-\system), and Common Ground (CG-MToMnet). 
Our goal is to explore whether an internal ToM mechanism can, similar to humans, also benefit computational agents.
We study different operations to combine neural representations for each variant of \system and identify the best for our tasks.

\paragraph{Decision-Based Theory of Mind (DB-\system).}
Inspired by previous work on referential games, where a speaker agent uses an internal listener model to re-rank potential utterances~\citep{liu2023computational}, DB-\system incorporates a ToM mechanism to re-rank class label predictions.
More specifically, we combine $P(y_i|\bm{x}_1)$ and $P(y_i|\bm{x}_2)$ within the \system using a ``ToM weight'' hyper-parameter $\tau$, and we take the argmax of this score as the final belief prediction:
\begin{align}
    \Tilde{b}_1 &= \argmax(P(y_i|\bm{x}_1)^\tau \cdot P(y_i|\bm{x}_2)) \\
    \Tilde{b}_2 &= \argmax(P(y_i|\bm{x}_2)^\tau \cdot P(y_i|\bm{x}_1))
\end{align}
In contrast to \citet{liu2023computational}, we apply the weight $\tau$ to the original probability distribution, e.g.\ to $P(y_i|\bm{x}_1)$ for MindNet$_1$, and not to the other probability distribution.
As such, the hyper-parameter $\tau$ controls the extent to which the prediction from one MindNet impacts that of the other:
the larger $\tau$, the smaller the impact. 

\paragraph{Implicit Communication Theory of Mind (IC-\system).} 
Conceptually similar to DB-\system, IC-\system enables communication between the two MindNets via internal representations instead of exchanging ranking scores.
Specifically, given the LSTM outputs
\begin{align}
    \bm{h}_1, \bm{c}_1 &= \texttt{LSTM}_1(\bm{x}_1) \qquad \bm{h}_2, \bm{c}_2 = \texttt{LSTM}_2(\bm{x}_2)
\end{align}
where $\bm{h}$ and $\bm{c}$ indicate the LSTM hidden state and cell state, we aggregate one MindNet's hidden state with the other MindNet's cell state, and vice versa. 
As we use bidirectional LSTMs, we use a fully connected layer to project the hidden state to the cell state dimension:
\begin{align}
    \bm{z}_1 &= \texttt{FC}(\bm{h}_1) \star \bm{c}_2 \qquad \bm{z}_2 = \texttt{FC}(\bm{h}_2) \star \bm{c}_1
\end{align}
where $\star$ can be one of the following aggregation operations: addition, multiplication, concatenation, or cross-attention~\citep{vaswani2017attention}.  
$\bm{z}_1$ and $\bm{z}_2$ are used to obtain the predictions in the final classification layers.

\paragraph{Common Ground Theory of Mind (CG-\system).}
This final variant is inspired by the idea that the human communicative context is not limited to the surrounding environment but 
involves a wider, shared and inter-subjective context known as common ground~\citep{clark1996using, tomasello2010origins}.
\citet{tomasello2010origins} refers to ``common ground'' as \textit{shared experience between individuals} that is critical for all human communication.
As such, common ground represents a broad concept that may include perception, attention, and knowledge. 
In this work, we build such common ground by combining the LSTM cell state $\bm{c}$ of each MindNet. 
We chose the LSTM cell state as it represents the memory of the network, storing information over time.
In practice, we concatenate the cell state of the two LSTMs to form a \textit{common ground representation}: 
\begin{align}
    \bm{cg} = \texttt{FC}(\bm{c}_1 \parallel \bm{c}_2) 
\end{align}
The $\bm{cg}$ tensor is then aggregated with the LSTM hidden states to obtain $\bm{z}_1$ and $\bm{z}_2$:
\begin{align}
    \bm{z}_1 &= \texttt{FC}(\bm{h}_1) \star \bm{cg} \qquad \bm{z}_2 = \texttt{FC}(\bm{h}_2) \star \bm{cg}
\end{align}
where $\star$ has the same meaning as before. 
$\bm{z}_1$ and $\bm{z}_2$ are used to obtain the predictions in the final classification layers.

\section{Experiments}

\subsection{Datasets}
\begin{figure}[t]
    \centering
    \includegraphics[width=\columnwidth]{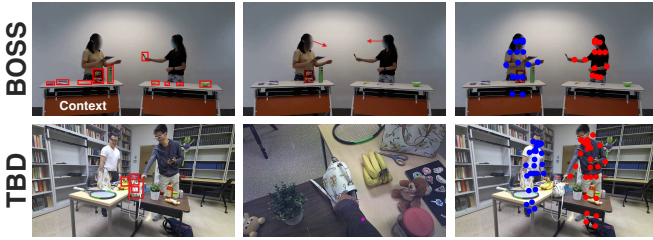}
    \caption{Examples from BOSS~\citep{duan2022boss} and TBD~\citep{fan2021learning}. BOSS includes third-person video frames, bounding boxes (top left), 3D gaze (top centre) and body pose (top right). TBD includes third-person (bottom left) and first-person (bottom centre) video frames, 2D gaze (bottom centre, pink dot) and body pose (bottom right).  
    }
    \vspace{0.7cm}
    \label{fig:datasets}
\end{figure}

\paragraph{BOSS.}
The Benchmark for Human Belief Prediction in Object-context Scenarios (BOSS) is a real-world dataset consisting of videos of two humans performing collaborative tasks without verbal communication~\citep{duan2022boss}. 
The dataset consists of 900 third-person videos recorded with 10 participants in 15 different object-context situations. 
In each video, each person stands in front of a table, one with contextual objects and the other with objects that can be selected based on the context (see Figure~\ref{fig:datasets}, top). 
The person standing in front of the contextual objects table receives a contextual task and must non-verbally communicate which object on the other table should be selected.
In Figure~\ref{fig:datasets} (top), the task is \textit{``Circle the words on the magazine's cover''}, for which the correct object to be selected by the participant on the right is the marker.

The computational task is to predict each person's beliefs for each video frame correctly.
This translates into a classification problem where the possible classes $\{c_i\}_{i=0}^{N}$ are the different objects, with $N=27$. 
The dataset is annotated with gaze estimation, pose estimation, bounding boxes, ground truth beliefs, and an object-context relations matrix (OCR) that describes which objects are more likely to appear given a certain context. 
For instance, a hammer is more likely to be present when there are nails rather than when there is a pizza.
Given that we noticed that the original bounding box annotations were highly inaccurate during preliminary experiments, we opted for re-extracting them using YOLOv5~\citep{yolov5}.
\footnote{Link to code and improved annotations in the Appendix.}

\paragraph{Triadic Belief Dynamics Dataset (TBD).}
\citet{fan2021learning} have collected a dataset covering nonverbal communication in rich social interactions. 
Participants were not provided a detailed script but only with the type of nonverbal communication they could use.
The dataset consists of 88 videos recorded with 12 people in seven different scenarios.
It includes first- and third-person video frames and gaze, pose, and bounding box annotations (see Figure~\ref{fig:datasets}, bottom). 
We opted for also evaluating on TBD because it differs from and complements BOSS in two distinct ways.
First, people were asked to perform three types of nonverbal communication -- no communication, attention following, and joint attention -- that, while highly relevant for belief prediction, do not directly involve collaboration. 
Second, TBD differs from BOSS in that the task is not to predict a person's belief about objects in the scene itself but its dynamics, i.e.\ if and how this belief changes over time. 
Concretely, given video clips of five frames, a model has to classify the belief dynamics for a selected object in the scene into four classes: \textit{occur}, \textit{disappear}, \textit{update}, and \textit{null}.
Importantly, TBD involves predicting belief dynamics not only for first-order ($m^1, m^2$) but also second-order beliefs ($m^{12}, m^{21}$) -- i.e.\ beliefs over another person's beliefs -- and a common mind ($m^c$) that corresponds to a common ground between the two participants.

\begin{figure*}[t]
    \centering
    \includegraphics[width=\textwidth]{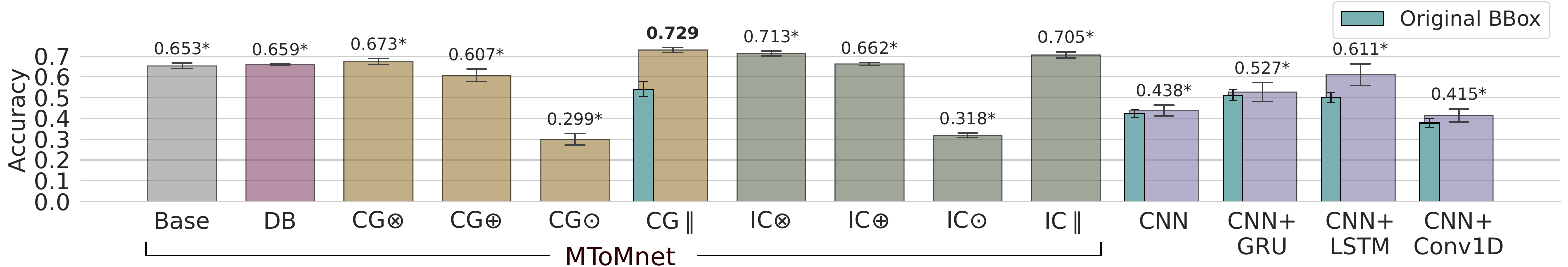}
    \vspace{0.01cm}
    \caption{Accuracy for belief prediction on BOSS for our \system models and baselines~\citep{duan2022boss} using all input modalities. 
    Scores significantly different from CG$\parallel$-\system according to a paired t-test ($p<0.05$) are marked with a *.
    }
    \vspace{0.4cm}
    \label{fig:boss_acc}
\end{figure*}

\subsection{Implementation Details}
\paragraph{\system.}
BOSS and TBD share most input modalities, with a few exceptions.
For BOSS, the contextual cues consist of third-person RGB videos, object bounding boxes, and the OCR matrix. 
Individual cues are derived from 3D gaze and pose.
TBD contains third-person RGB videos and object bounding boxes as contextual cues and first-person RGB videos, pose, and 2D gaze
as individual cues.
Given these differences, we implemented dataset-specific feature extractors. 
All \system variants encoded RGB video frames using a three-layer CNN with $16$, $32$, and $64$ filters, respectively.
Each convolutional layer was followed by ReLU activation and max pooling.
The OCR matrix and poses were processed by a graph convolutional network~\citep{kipf2017semi}.
The OCR matrix naturally represents an adjacency matrix for our graphs, and we used its normalised values as node features.
For the poses, we defined the adjacency matrix based on connections of body joints and used the 3D joint coordinates as node features.
Object bounding boxes are fed into a fully connected layer.
Each MindNet uses a one-layer bidirectional LSTM, preceded by layer normalisation. 
All layers in our \system models have hidden dimension $64$ and are followed by GELU activation~\citep{hendrycks2016gaussian} and dropout~\citep{srivastava2014dropout} with $p=0.1$. 
For DB-\system we set the ``ToM weight'' to $\tau=2$.
For BOSS, each MindNet outputs a belief prediction. 
For TBD, classification layers for $m^1$ and $m^{12}$ take $\bm{z}_1$ as input, whereas classification layers for $m^2$ and $m^{21}$ take $\bm{z}_2$ as input.
Since $m^c$ represents the ``common mind'' between both agents \cite{fan2021learning}, $\bm{z}_1$ and $\bm{z}_2$ are first aggregated by performing element-wise multiplication and then fed into the classification layer.
Additional details on the architecture are provided in Appendix.

\paragraph{Training.} 
We trained all models for $300$ epochs using three distinct random seeds. 
Cross-entropy was employed as the loss function, and model checkpoints were saved based on the highest validation accuracy for BOSS and the highest macro F1 score for TBD. 
These metrics were chosen for easier comparison with the original works \citep{duan2022boss,fan2021learning}. 
We used the Adam optimiser~\citep{kingma2015adam} with a learning rate of $5\cdot10^{-4}$.
Additional details are provided in Appendix.

\begin{table*}[t]
    \centering
    \caption{Macro F1 scores for previous approaches~\citep{fan2021learning} and \system variants on TBD. Scores significantly different from CG$\parallel$-\system according to a paired t-test ($p<0.05$) are marked with a *.}
    \resizebox{\textwidth}{!}{
    \begin{tabular}{ccccccc}
        \toprule
        \textbf{Model} & $\bm{m^1}$ & $\bm{m^2}$ & $\bm{m^{12}}$ & $\bm{m^{21}}$ & $\bm{m^c}$ & \textbf{Average} \\
        \midrule
        Chance & $0.103$ & $0.104$ & $0.102$ & $0.101$ & $0.100$ & $0.102$ \\
        CNN & $0.171$ & $0.167$ & $0.169$ & $0.174$ & $0.250$ & $0.186$ \\
        CNN+HOG-LSTM & $0.167$ & $0.132$ & $0.205$ & $0.182$ & $0.250$ & $0.187$ \\
        CNN+HOG+Mem & $0.285$ & $0.285$ & $0.246$ & $0.250$ & $0.155$ & $0.244$ \\
        Feats+Mem & $0.274$ & $0.288$ & $0.230$ & $0.227$ & $0.191$ & $0.242$ \\
        HGM & $0.431$ & $0.443$ & $0.351$ & $0.349$ & $0.299$ & $0.375$ \\
        \midrule
        -\system & & & & & & \\
        \cmidrule{1-1} 
        Base & $0.276 \pm 0.055$* & $0.473 \pm 0.085$* & $0.313 \pm 0.014$* & $0.473 \pm 0.010$ & $0.566 \pm 0.095$* & $0.420 \pm 0.023$ \\
        DB & $0.442 \pm 0.067$* & $0.313 \pm 0.051$* & $0.425 \pm 0.006$* & $0.385 \pm 0.013$* & $0.459 \pm 0.010$* & $0.405 \pm 0.013$ \\
        CG$\parallel$ & $\mathbf{0.477 \pm 0.078}$ & $0.460 \pm 0.065$ & $0.452 \pm 0.010$ & $0.467 \pm 0.013$ & $\mathbf{0.583 \pm 0.080}$ & $\mathbf{0.488 \pm 0.022}$ \\
        CG$\oplus$ & $0.461 \pm 0.059$ & $0.472 \pm 0.076$ & $0.459 \pm 0.009$ & $0.468 \pm 0.012$* & $0.544 \pm 0.091$ & $0.481 \pm 0.022$ \\
        CG$\otimes$ & $0.462 \pm 0.062$* & $\mathbf{0.478 \pm 0.068}$ & $0.451 \pm 0.010$ & $0.469 \pm 0.014$* & $0.559 \pm 0.103$* & $0.484 \pm 0.023$ \\
        CG$\odot$ & $0.461 \pm 0.057$ & $0.464 \pm 0.062$* & $\mathbf{0.469 \pm 0.015}$ & $0.475 \pm 0.009$ & $0.564 \pm 0.099$ & $0.486 \pm 0.022$ \\
        IC$\parallel$ & $0.462 \pm 0.067$* & $0.463 \pm 0.067$ & $0.459 \pm 0.002$ & $0.471 \pm 0.014$ & $0.556 \pm 0.093$* & $0.482 \pm 0.022$ \\
        IC$\oplus$ & $0.465 \pm 0.074$* & $0.474 \pm 0.073$* & $0.466 \pm 0.025$* & $0.473 \pm 0.015$ & $0.561 \pm 0.098$ & $\mathbf{0.488 \pm 0.025}$ \\
        IC$\otimes$ & $0.417 \pm 0.066$* & $0.423 \pm 0.072$ & $0.464 \pm 0.001$ & $\mathbf{0.486 \pm 0.007}$ & $0.450 \pm 0.007$* & $0.448 \pm 0.014$ \\
        IC$\odot$ & $0.455 \pm 0.072$* & $0.257 \pm 0.002$* & $0.462 \pm 0.032$* & $0.274 \pm 0.016$* & $0.554 \pm 0.084$* & $0.401 \pm 0.019$ \\
        \bottomrule
    \end{tabular}
    }
    \label{tab:tbd_f1}
\end{table*}

\paragraph{Baselines.} 
We compare our approach with the original models~\citep{fan2021learning, duan2022boss}.
For BOSS, \citet{duan2022boss} have proposed four models based on a ResNet34 backbone for encoding video frames and fully connected layers for other modalities like gaze, pose, bounding boxes, and OCR. 
The models are CNN, CNN+GRU, CNN+LSTM, and CNN+Conv1D, each incorporating different types of recurrent or convolutional layers before passing the concatenated latent representations to two classification layers.
To ensure a fair comparison, we re-trained these models for $300$ epochs, as the original models were only trained for five epochs.

\citet{fan2021learning} have evaluated similar approaches on TBD.
Their first model (CNN) uses a ResNet50 to extract frame features, followed by fully connected layers for belief dynamics classification. 
The CNN+HOG-LSTM model incorporates histograms of oriented gradient~\citep[HOG]{dalal2005histograms} features of the frame patch gazed at by the participants along with full frame features. 
The CNN+HOG+Mem model concatenates the history of predicted belief dynamics with frame and HOG features, while the Feats+Memory model combines handcrafted features with the history of predicted belief dynamics using a multi-layer perceptron.
\citet{fan2021learning} achieved state-of-the-art performance on TBD by using a hierarchical graphical model (denoted here as HGM)
trained using a beam-search algorithm on handcrafted events derived from raw pixels.

\subsection{Model Performance}
Results for the different \system variants and the baselines on BOSS are shown in Figure~\ref{fig:boss_acc}. 
We use the following notation to indicate the different aggregation operations: $\otimes$ = element-wise multiplication, $\oplus$ = element-wise sum, $\parallel$ = concatenation, $\odot$ = self-attention.
As can be seen from the figure,
already the Base-\system (i.e.\ without explicit ToM modelling) outperforms all baselines ($0.653$ accuracy), despite requiring less than $3\%$ of their parameters ($\sim450$k vs $21$M).
Second, incorporating explicit ToM modelling (DB, CG, IC) yields further performance improvements, but the choice of aggregation is critical. 
CG$\parallel$-\system exhibits the highest overall performance amongst all models ($0.729$), followed by IC$\otimes$-\system ($0.713$), and IC$\parallel$-\system ($0.705$).
In contrast, the DB-\system ($0.659$) only achieves a marginal improvement over the Base-\system ($0.653$).
Figure~\ref{fig:boss_acc} also shows results for CG$\parallel$-\system and baselines obtained using the original bounding box annotations. 
While CG$\parallel$-\system still outperforms all the baselines, the baselines do not benefit from the improved bounding boxes. This is likely attributed to the shallow feature aggregation in the baseline models. 
The paired t-test revealed a significant difference ($p<0.05$) between CG$\parallel$-\system and the other models. 
To further validate our architectural choice, we evaluated a single MindNet model. 
Our methods outperform this model and achieve an accuracy of $0.61$, except for CG$\odot$ and IC$\odot$.
We report modality ablation studies in Appendix.

Evaluation scores for belief dynamics prediction on TBD are shown in Table~\ref{tab:tbd_f1}, where we also report paired t-test results ($p<0.05$) between CG$\parallel$-\system and other \system variants.
As for BOSS, our Base-\system already achieves better performance than the best baseline (HGM) -- with the only exception of $m^1$ ($0.276$ vs.\ $0.431$) and $m^{12}$ ($0.313$ vs.\ $0.351$). 
Adding explicit ToM modelling leads to further performance gains for all three variants, up to $30\%$ on HGM. 
Considering the average performance across different aggregation types, the best model is again the one inspired by social cognition, CG-\system.
In particular, the single best performing models are CG$\parallel$-\system and IC$\oplus$-\system, on par ($0.488$).
Remarkably, all our \system variants achieved their highest F1 scores on $m^c$, except for IC$\otimes$-\system. 
In contrast, baselines typically find classifying belief dynamics for $m^c$ one of the most challenging tasks, achieving lower scores. 
Our CG$\parallel$-\system ($0.583$) substantially outperformed the best baseline, HGM ($0.299$), by a substantial margin of improvement. 

These results highlight the effectiveness of our proposed \system architecture and underline the significance of explicit ToM modelling in achieving superior performance with significantly reduced computational costs.

\paragraph{Modality Ablation Study.}
\begin{figure}[t]
    \centering
    \includegraphics[width=\columnwidth]{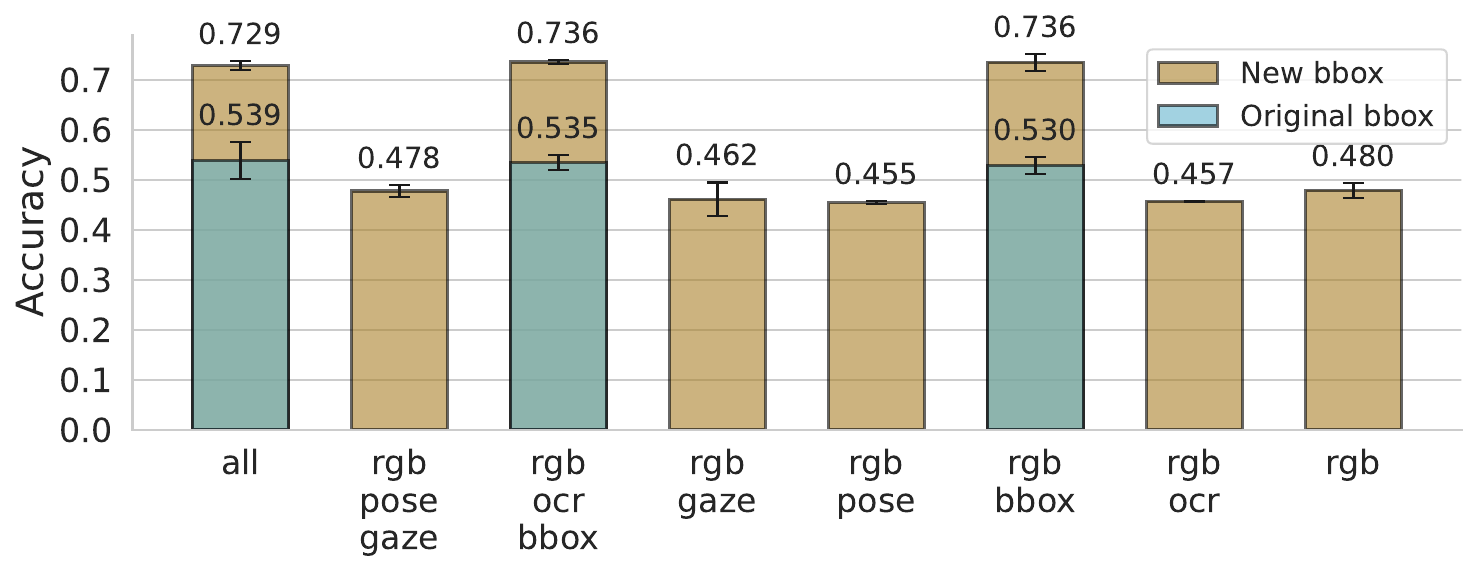}
    \caption{Modality ablation study for BOSS.}
    \label{fig:boss_abl}
    \vspace{0.5cm}
\end{figure}

\begin{figure}[t]
    \centering
    \includegraphics[width=\columnwidth]{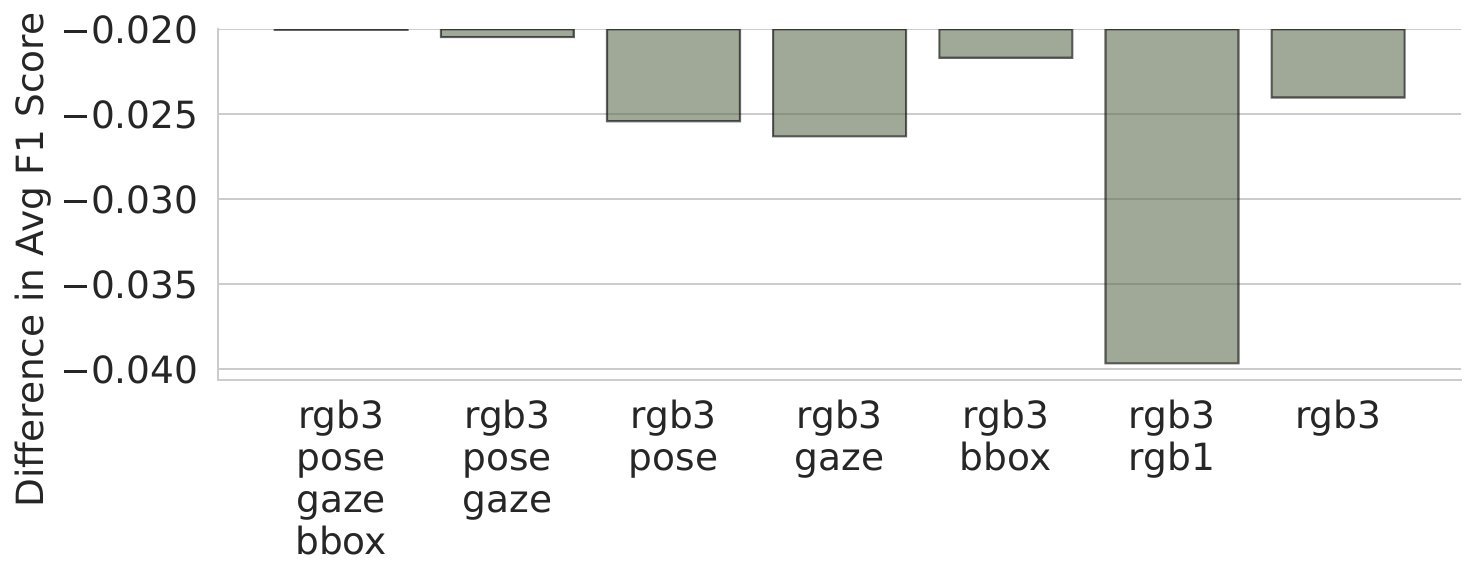}
    \caption{Modality ablation study for TBD.}
    \label{fig:tbd_abl}
    \vspace{0.5cm}
\end{figure}

Figure~\ref{fig:boss_abl} shows the accuracy achieved by ablated versions of our best-performing model, CG$\parallel$-\system, on the BOSS dataset. 
The results highlight the significant impact of including bounding boxes as input modality (rgb+ocr+bbox and rgb+bbox).
This result aligns with previous research~\citep{duan2022boss}, emphasising the crucial role of knowing the objects present in the scene for the task.
When excluding bounding boxes, the accuracy scores decreased. 
We suspected that the newly introduced bounding box annotations might be a major contributing factor to this outcome. 
Therefore, for the sake of completeness, we conducted additional experiments where CG$\parallel$-\system was trained and evaluated using the original bounding box data.
The results, depicted in Figure~\ref{fig:boss_abl} (teal), demonstrate that when utilising the original bounding boxes, the performance gap with other ablated versions of the model drastically decreases. 
Nevertheless, our model performs better than the baselines even when using original bounding box annotations.

The differences in averaged F1 scores across $m^i$, $i=\{1,2,12,21\}$ on TBD between the complete CG$\parallel$-\system and its ablated versions are reported in Figure~\ref{fig:tbd_abl}.
Our full model achieves an F1 score of $0.488$. 
Ablating modalities generally result in performance degradation, with the worst-performing version experiencing an 8.9\% decline, attaining an F1 score of $0.449$ (rgb1+rgb3). 
Integrating multiple behavioural cues, such as pose and gaze, contributes positively to performance. 
Specifically, the combination of third-person RGB frames with gaze and pose achieved an F1 score of $0.487$, outperforming models where third-person RGB frames were combined with only gaze or pose.

\begin{figure}[t]
    \centering
    \begin{subfigure}{.47\columnwidth}
        \centering
        \includegraphics[width=\linewidth]{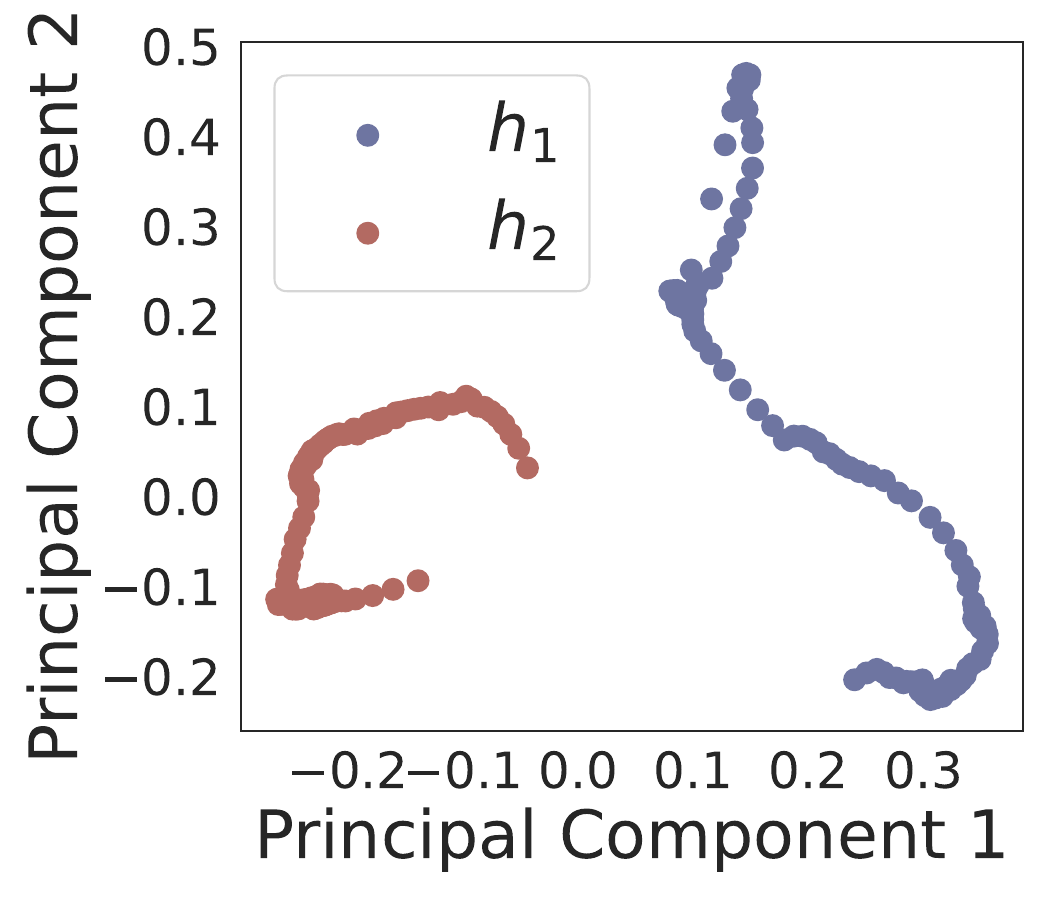}
        \caption{BOSS}
        \label{fig:pca_boss}
    \end{subfigure}%
    \hfill
    \begin{subfigure}{.46\columnwidth}
        \centering
        \includegraphics[width=\linewidth]{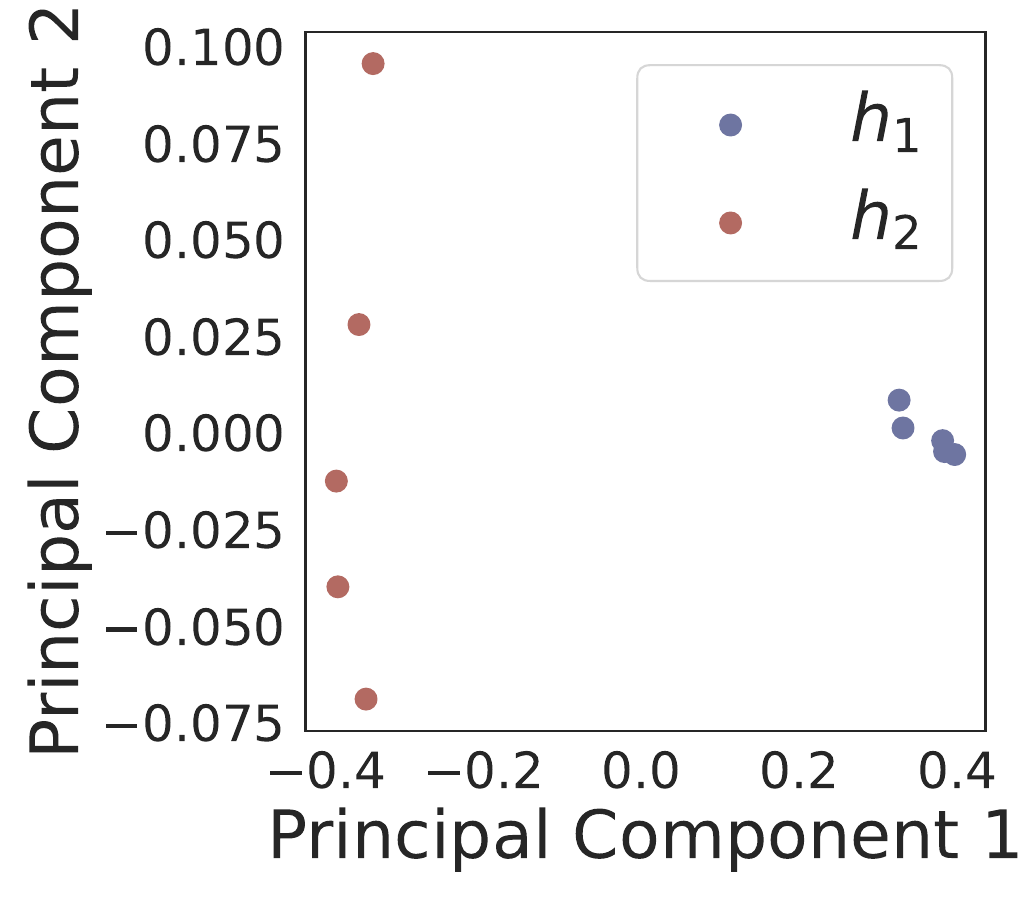}
        \caption{TBD}
        \label{fig:pca_tbd}
    \end{subfigure}
    \vspace{0.4cm}
    \caption{Examples from PCA results for $\bm{h}_1$ and $\bm{h}_2$ from CG$\parallel$-\system, taken from BOSS and TBD test set.}
    \vspace{0.7cm}
    \label{fig:pca}
\end{figure}

\begin{figure*}
    \centering
    \includegraphics[width=\textwidth]{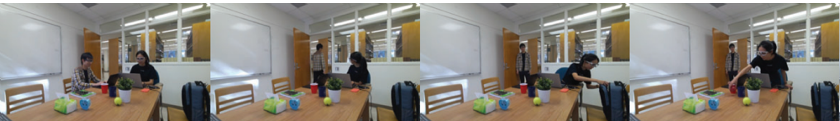}
    \caption{Example of \textit{second-order} false belief from TBD \cite{fan2021learning}. Thinking her partner had left the room, a participant moved an apple from the backpack to behind her laptop. \textit{She believes her partner believes} the apple is still in the backpack. However, she is unaware that her partner is actually standing by the door and saw her move the apple to the table.}
    \label{fig:tbd-fb-example}
\end{figure*}

\begin{figure*}[t]
    \centering
    \includegraphics[width=\textwidth]{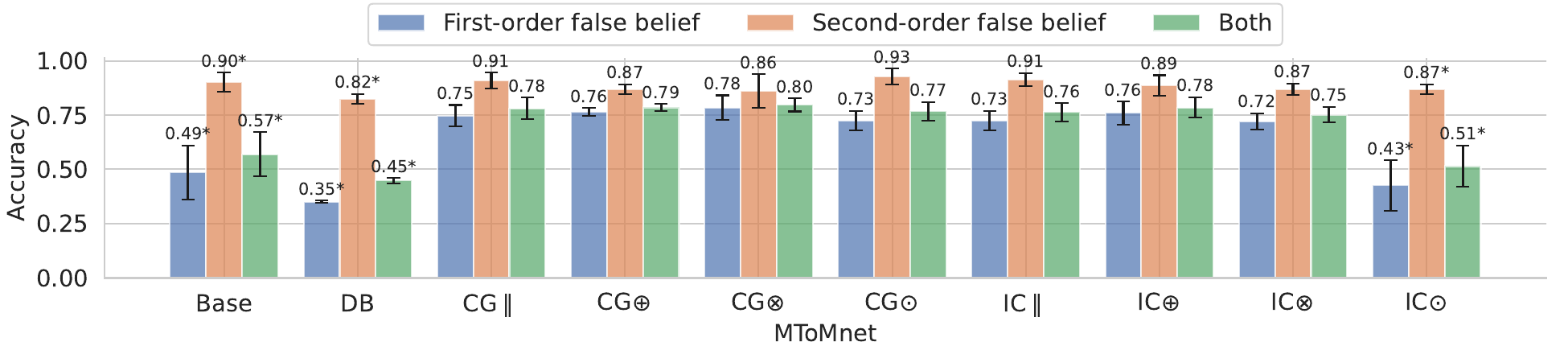}
    \caption{Accuracy of false belief prediction on TBD for different variants of our \system (Base = no explicit ToM modelling, DB = decision-based ToM, CG = common ground ToM, IC = implicit communication ToM). Scores significantly different from CG$\parallel$-\system according to a paired t-test ($p<0.05$) are marked with a *.}
    \vspace{0.1cm}
    \label{fig:tbd_false_belief_first_vs_second}
\end{figure*}

\subsection{Modelling Person-Specific Features}

To assess the efficacy of encoding individual cues and predicting individuals' beliefs using independent MindNets, we compared the feature representations from the two LSTMs, $\bm{h}_1$ and $\bm{h}_2$, by employing Principal Component Analysis~\citep[PCA]{pearson1901liii}. 
As the examples in Figure~\ref{fig:pca} show, the LSTMs hidden states $\bm{h}_1$ and $\bm{h}_2$ from CG$\parallel$-\system show a clear disentanglement of principal components both on BOSS and TBD.
We found the same behaviour for all our \system variants and provide examples for all of them in Appendix. 
This finding suggests that each MindNet represents person-specific cues and fuses them with contextual cues differently for different persons, highlighting the value of such architectural choice.

\subsection{False Belief Dynamics Prediction}

\begin{table}[t]
    \centering
    \caption{Label counts for TBD train and test set.}
    \resizebox{\columnwidth}{!}{
    \begin{tabular}{lccccc}
        \toprule
        \multirow{2}{*}{\textbf{Mind}} & \multicolumn{4}{c}{\textbf{Train/test count}} \\
        \cmidrule{2-5} & Occur & Disappear & Update & Null \\
        \midrule
        $m^1$ (all) & $156/48$ & $0/0$ & $2176/731$ & $1750/582$ \\
        $m^1$ (false belief) & $0/1$ & $0/0$ & $5/0$ & $158/53$ \\
        $m^2$ (all) & $146/46$ & $9/0$ & $2579/860$ & $1348/455$ \\
        $m^2$ (false belief) & $0/1$ & $0/0$ & $1/1$ & $118/47$ \\
        $m^{12}$ (all) & $75/25$ & $16/5$ & $916/298$ & $3075/1033$ \\
        $m^{12}$ (false belief) & $0/0$ & $0/0$ & $0/0$ & $19/4$ \\
        $m^{21}$ (all) & $81/27$ & $11/1$ & $821/260$ & $3169/1073$ \\
        $m^{21}$ (false belief) & $0/0$ & $0/0$ & $0/0$ & $52/20$ \\
        \bottomrule
    \end{tabular}
    }
    \label{tab:label_counts}
\end{table}

TBD involves predicting belief dynamics for both first- and second-order beliefs, i.e., self-beliefs $(m^1, m^2)$ and beliefs held over another person's beliefs $(m^{12}, m^{21})$.
Human beliefs, however, are not always aligned with reality: individuals may hold a first- or second-order \textit{false belief}~\citep{wimmer1983beliefs}. 
We show an example in Figure~\ref{fig:tbd-fb-example}. Thinking her partner had left the room, a participant moved an apple from the backpack to behind her laptop. 
She believes her partner believes the apple is still in the backpack. 
However, she is unaware that her partner is actually standing by the door and saw her move the apple to the table. 
This situation exemplifies a second-order (\textit{she believes he thinks}) false belief.
Despite not being used in~\citep{fan2021learning}, the TBD dataset includes false belief annotations, allowing us to perform post-hoc analyses on models' capabilities to predict such false belief dynamics.
That is, 
we do not train models specifically to recognise false beliefs but evaluate models' belief dynamics predictions that, according to the provided annotations, correspond to false beliefs. 

Figure~\ref{fig:tbd_false_belief_first_vs_second} summarises the accuracy for false belief dynamics predictions on TBD, considering first-order, second-order, and both false belief types. 
We report accuracy as we were interested in detecting whether a model predicted correctly 
(positive class) 
or not 
(negative class) 
a 
(a certain type of) 
false belief.
For first-order false belief and joint first- and second-order false belief dynamics prediction, all IC- and CG-\system variants outperform Base, except for IC$\odot$. 
In particular, our CG-\system variants achieve the highest accuracy, improving over Base-\system by a large margin on first-order false beliefs (up to $0.78$ vs.\ $0.49$) and on joint first- and second-order false beliefs (up to $0.80$ vs.\ $0.57$).

Figure~\ref{fig:tbd_false_belief_first_vs_second} also shows that predicting second-order false belief dynamics -- generally more difficult -- is easier on TBD. 
To understand why, we compared the distribution of all labels with those corresponding to false beliefs, as shown in Table~\ref{tab:label_counts}.
In training and test sets, most false belief labels for all minds $m^i$, $i=\{1,2,12,21\}$, are \textit{null}. 
For false beliefs associated with $m^{12}$ and $m^{21}$, \textit{null} is the only possible label.
Thus, most false beliefs in the dataset correspond to situations where individuals assume that nothing changed although something did (\textit{occur}, \textit{disappear}, or \textit{update}) -- akin to the Sally-Anne test~\citep{baron1985does}.
However, when considering the overall label distribution, the most frequent label for $m^1$ and $m^2$ is \textit{update}, thus leading to biases towards predicting the \textit{update} class during training. 
This ultimately leads to lower accuracy in predicting first-order false beliefs, where \textit{null} is the most prevalent label.
However, our \system variants, especially CG-\system, can overcome this bias.

\section{Discussion}

\vspace{-0.2cm}

\paragraph{Social Cognition Is All You Need.}
In this work, we presented \system~-- a Theory of Mind neural network for predicting beliefs and their dynamics during nonverbal human social interactions.
Our best performing \system variant (\textbf{CG-\system}) \textbf{achieved new state-of-the-art results} for belief (dynamics) prediction on both BOSS (Figure~\ref{fig:boss_acc}) and TBD (Table~\ref{tab:tbd_f1}), as well as for false belief dynamics prediction on TBD (Figure~\ref{fig:tbd_false_belief_first_vs_second}). 
At the same time, \textbf{\system requires considerably fewer parameters than previous state-of-the-art methods} -- approximately $460$k vs.\ $21$M for BOSS and $1$M for TBD (see Appendix) -- and, as a result, shows faster training (less than 3 hours vs. 17 to 20 hours for baselines on BOSS~\citep{duan2022boss}). 
These findings are important as they underline the significant potential of adopting concepts developed in the cognitive sciences when designing computational agents.
This also represents a paradigm shift compared to 
the recent trend of improving performance mainly through upscaling model complexity.
Our analyses of the latent representations showed that the two MindNets are effective in capturing individuals' information in distinct ways for modelling beliefs (Figure~\ref{fig:pca}), further supporting the architectural design choices of \system.
Remarkably, \textbf{CG-\system can be easily adapted to interactions involving more than two interacting agents}, thanks to its shared common ground representation across all MindNets. 
This adaptability is crucial for future work exploring scenarios with several human and computational agents.
Extending DB- and IC-\system to involve more than two agents is more challenging. 
For DB-\system, it would be crucial to find the right set of $\tau$ such that probabilities do not vanish. 
One interesting idea for future work is to make $\tau$ learnable. 
IC$\parallel$-\system faces the challenge of aggregating numerous hidden states, which could result in a quite large vector. 
A broader limitation arises when dealing with dynamic environments, where the number of individuals in a scene can vary.
Together with extending models to dynamic environments, another promising direction is to improve the integration of different input features, a facet not explored in this work. 
For example, on BOSS, the OCR matrix could be used to define an additional term in the loss function to enforce object-context relations.

\paragraph{Explicit Modelling of ToM Improves Performance.}
A key contribution of our work is the explicit modelling of ToM using multimodal individual and contextual cues.
Our experiments demonstrated that \system not only achieves new state-of-the-art performance on the two most common benchmark datasets but also outperforms existing baselines by a large margin.
Instrumental to these improvements is the explicit ToM modelling that allowed us to improve our Base-\system 
by $19\%$ on BOSS (CG$\parallel$-\system, Figure~\ref{fig:boss_acc}), and up to $60\%$ on TBD (CG$\parallel$-\system, Table~\ref{tab:tbd_f1}, Average).
This particularly shows for the challenging task of false belief prediction on TBD for which explicit ToM modelling leads to significant improvements of up to $60\%$ over the Base-\system model (Figure~\ref{fig:tbd_false_belief_first_vs_second}).
This finding is particularly significant considering the large body of work and long-standing efforts on false belief prediction \citep{goodman2006intuitive, baker2017rational, rabinowitz2018machine, ullman2023large}.

\paragraph{Limitations of Current Benchmarks.}
Our ablation studies highlight a fundamental limitation inherent in current benchmarks due to the quality and consistency of the data, negatively affecting model performance. 
In this work, we found inaccuracies in bounding box annotations in the BOSS dataset and re-extracted them. 
This led to a substantial enhancement in performance, highlighting the importance of precise bounding box information. 
Rectifying similar issues in gaze data proved unfeasible, as participant faces in the BOSS dataset were deliberately obscured for privacy reasons.
Despite data limitations, our models outperformed the baselines with both original and revised annotations (Figure~\ref{fig:boss_abl}).

\section{Conclusion}

In this work, we proposed \system, a Theory of Mind network that predicts beliefs and their dynamics during human social interactions from multimodal input. 
Building on social cognition and ToM research, we designed three \system variations: one decision-based and two model-based.
Across two real-world datasets, \system outperformed existing methods in both belief prediction and belief dynamics prediction, despite having fewer parameters. 
These results advance the state-of-the-art in belief prediction thus facilitating better collaboration with humans.

\section*{Acknowledgements}
M. Bortoletto and A. Bulling were funded by the European Research Council (ERC) under the grant agreement 801708.
L. Shi was funded by the Deutsche Forschungsgemeinschaft (DFG, German Research Foundation) under Germany’s Excellence Strategy -- EXC 2075 -- 390740016.
The authors thank the International Max Planck Research School for Intelligent Systems (IMPRS-IS) for supporting C. Ruhdorfer.

\bibliography{refs}

\begin{thebibliography}{49}
\providecommand{\natexlab}[1]{#1}
\providecommand{\url}[1]{\texttt{#1}}
\expandafter\ifx\csname urlstyle\endcsname\relax
  \providecommand{\doi}[1]{doi: #1}\else
  \providecommand{\doi}{doi: \begingroup \urlstyle{rm}\Url}\fi

\bibitem[Ba et~al.(2016)Ba, Kiros, and Hinton]{ba2016layer}
J.~L. Ba, J.~R. Kiros, and G.~E. Hinton.
\newblock Layer normalization.
\newblock \emph{arXiv preprint arXiv:1607.06450}, 2016.

\bibitem[Baker et~al.(2009)Baker, Saxe, and Tenenbaum]{baker2009action}
C.~L. Baker, R.~Saxe, and J.~B. Tenenbaum.
\newblock Action understanding as inverse planning.
\newblock \emph{Cognition}, 113\penalty0 (3):\penalty0 329--349, 2009.

\bibitem[Baker et~al.(2017)Baker, Jara-Ettinger, Saxe, and
  Tenenbaum]{baker2017rational}
C.~L. Baker, J.~Jara-Ettinger, R.~Saxe, and J.~B. Tenenbaum.
\newblock Rational quantitative attribution of beliefs, desires and percepts in
  human mentalizing.
\newblock \emph{Nature Human Behaviour}, 1\penalty0 (4):\penalty0 0064, 2017.

\bibitem[Bara et~al.(2021)Bara, CH-Wang, and Chai]{bara2021mindcraft}
C.-P. Bara, S.~CH-Wang, and J.~Chai.
\newblock {M}ind{C}raft: Theory of mind modeling for situated dialogue in
  collaborative tasks.
\newblock In \emph{Proceedings of the 2021 Conference on Empirical Methods in
  Natural Language Processing}, pages 1112--1125. Association for Computational
  Linguistics, 2021.
\newblock \doi{10.18653/v1/2021.emnlp-main.85}.

\bibitem[Baron-Cohen et~al.(1985)Baron-Cohen, Leslie, and Frith]{baron1985does}
S.~Baron-Cohen, A.~M. Leslie, and U.~Frith.
\newblock Does the autistic child have a “theory of mind”?
\newblock \emph{Cognition}, 21\penalty0 (1):\penalty0 37--46, 1985.

\bibitem[Bortoletto et~al.(2024{\natexlab{a}})Bortoletto, Ruhdorfer,
  Abdessaied, Shi, and Bulling]{bortoletto24Limits}
M.~Bortoletto, C.~Ruhdorfer, A.~Abdessaied, L.~Shi, and A.~Bulling.
\newblock Limits of theory of mind modelling in dialogue-based collaborative
  plan acquisition.
\newblock In \emph{Proc. 62nd Annual Meeting of the Association for
  Computational Linguistics (ACL)}, pages 1--16, 2024{\natexlab{a}}.

\bibitem[Bortoletto et~al.(2024{\natexlab{b}})Bortoletto, Ruhdorfer, Shi, and
  Bulling]{bortoletto2024benchmarking}
M.~Bortoletto, C.~Ruhdorfer, L.~Shi, and A.~Bulling.
\newblock Benchmarking mental state representations in language models.
\newblock In \emph{ICML 2024 Workshop on Mechanistic Interpretability},
  2024{\natexlab{b}}.
\newblock URL \url{https://openreview.net/forum?id=yEwEVoH9Be}.

\bibitem[Bortoletto et~al.(2024{\natexlab{c}})Bortoletto, Shi, and
  Bulling]{bortoletto2024neural}
M.~Bortoletto, L.~Shi, and A.~Bulling.
\newblock Neural reasoning about agents’ goals, preferences, and actions.
\newblock In \emph{Proceedings of the AAAI Conference on Artificial
  Intelligence}, volume~38, pages 456--464, 2024{\natexlab{c}}.

\bibitem[Clark(1996)]{clark1996using}
H.~H. Clark.
\newblock \emph{Using language}.
\newblock Cambridge University Press, 1996.

\bibitem[Dalal and Triggs(2005)]{dalal2005histograms}
N.~Dalal and B.~Triggs.
\newblock Histograms of oriented gradients for human detection.
\newblock In \emph{Proceedings of the IEEE/CVF Conference on Computer Vision
  and Pattern Recognition}, volume~1, pages 886--893. IEEE, 2005.

\bibitem[Doshi et~al.(2010)Doshi, Qu, Goodie, and Young]{doshi2010modeling}
P.~Doshi, X.~Qu, A.~Goodie, and D.~Young.
\newblock Modeling recursive reasoning by humans using empirically informed
  interactive pomdps.
\newblock In \emph{Proceedings of the International Conference on Autonomous
  Agents and Multiagent Systems}, pages 1223--1230, 2010.

\bibitem[Duan et~al.(2022)Duan, Yu, Tan, Yi, and Tan]{duan2022boss}
J.~Duan, S.~Yu, N.~Tan, L.~Yi, and C.~Tan.
\newblock Boss: A benchmark for human belief prediction in object-context
  scenarios.
\newblock \emph{arXiv preprint arXiv:2206.10665}, 2022.

\bibitem[Fan et~al.(2021)Fan, Qiu, Zheng, Gao, Zhu, and Zhu]{fan2021learning}
L.~Fan, S.~Qiu, Z.~Zheng, T.~Gao, S.-C. Zhu, and Y.~Zhu.
\newblock Learning triadic belief dynamics in nonverbal communication from
  videos.
\newblock In \emph{Proceedings of the IEEE/CVF Conference on Computer Vision
  and Pattern Recognition}, pages 7312--7321, 2021.

\bibitem[Floyd(2011)]{floyd2011interpersonal}
K.~Floyd.
\newblock \emph{Interpersonal communication}.
\newblock McGraw-Hill, 2011.

\bibitem[Gandhi et~al.(2021)Gandhi, Stojnic, Lake, and Dillon]{gandhi2021baby}
K.~Gandhi, G.~Stojnic, B.~M. Lake, and M.~R. Dillon.
\newblock Baby intuitions benchmark (bib): Discerning the goals, preferences,
  and actions of others.
\newblock \emph{Advances in Neural Information Processing Systems},
  34:\penalty0 9963--9976, 2021.

\bibitem[Goodman et~al.(2006)Goodman, Baker, Bonawitz, Mansinghka, Gopnik,
  Wellman, Schulz, and Tenenbaum]{goodman2006intuitive}
N.~D. Goodman, C.~L. Baker, E.~B. Bonawitz, V.~K. Mansinghka, A.~Gopnik,
  H.~Wellman, L.~Schulz, and J.~B. Tenenbaum.
\newblock Intuitive theories of mind: A rational approach to false belief.
\newblock In \emph{Proceedings of the twenty-eighth annual conference of the
  cognitive science society}, volume~6. Cognitive Science Society Vancouver,
  2006.

\bibitem[Graves and Schmidhuber(2005)]{graves2005framewise}
A.~Graves and J.~Schmidhuber.
\newblock Framewise phoneme classification with bidirectional lstm networks.
\newblock In \emph{Proceedings of the IEEE International Joint Conference on
  Neural Networks}, volume~4. IEEE, 2005.

\bibitem[Gurney and Pynadath(2022)]{gurney2022robots}
N.~Gurney and D.~V. Pynadath.
\newblock Robots with theory of mind for humans: {A} survey.
\newblock In \emph{Proceedings of the IEEE International Conference on Robot
  and Human Interactive Communication}. IEEE, 2022.

\bibitem[Han and Gmytrasiewicz(2018)]{han2018learning}
Y.~Han and P.~Gmytrasiewicz.
\newblock Learning others' intentional models in multi-agent settings using
  interactive pomdps.
\newblock \emph{Advances in Neural Information Processing Systems}, 31, 2018.

\bibitem[Harris et~al.(2020)Harris, Millman, van~der Walt, Gommers, Virtanen,
  Cournapeau, Wieser, Taylor, Berg, Smith, Kern, Picus, Hoyer, van Kerkwijk,
  Brett, Haldane, del R{\'{i}}o, Wiebe, Peterson, G{\'{e}}rard-Marchant,
  Sheppard, Reddy, Weckesser, Abbasi, Gohlke, and Oliphant]{harris2020array}
C.~R. Harris, K.~J. Millman, S.~J. van~der Walt, R.~Gommers, P.~Virtanen,
  D.~Cournapeau, E.~Wieser, J.~Taylor, S.~Berg, N.~J. Smith, R.~Kern, M.~Picus,
  S.~Hoyer, M.~H. van Kerkwijk, M.~Brett, A.~Haldane, J.~F. del R{\'{i}}o,
  M.~Wiebe, P.~Peterson, P.~G{\'{e}}rard-Marchant, K.~Sheppard, T.~Reddy,
  W.~Weckesser, H.~Abbasi, C.~Gohlke, and T.~E. Oliphant.
\newblock Array programming with {NumPy}.
\newblock \emph{Nature}, 585\penalty0 (7825):\penalty0 357--362, Sept. 2020.
\newblock \doi{10.1038/s41586-020-2649-2}.
\newblock URL \url{https://doi.org/10.1038/s41586-020-2649-2}.

\bibitem[He et~al.(2016)He, Zhang, Ren, and Sun]{he2016deep}
K.~He, X.~Zhang, S.~Ren, and J.~Sun.
\newblock Deep residual learning for image recognition.
\newblock In \emph{Proceedings of the IEEE/CVF Conference on Computer Vision
  and Pattern Recognition}, pages 770--778, 2016.

\bibitem[Hendrycks and Gimpel(2016)]{hendrycks2016gaussian}
D.~Hendrycks and K.~Gimpel.
\newblock Gaussian error linear units (gelus).
\newblock \emph{arXiv preprint arXiv:1606.08415}, 2016.

\bibitem[Hunter(2007)]{Hunter2007matplotlib}
J.~D. Hunter.
\newblock Matplotlib: A 2d graphics environment.
\newblock \emph{Computing in Science \& Engineering}, 9\penalty0 (3):\penalty0
  90--95, 2007.
\newblock \doi{10.1109/MCSE.2007.55}.

\bibitem[Jocher(2020)]{yolov5}
G.~Jocher.
\newblock Yolov5 by ultralytics, 2020.
\newblock URL \url{https://github.com/ultralytics/yolov5}.

\bibitem[Kingma and Ba(2015)]{kingma2015adam}
D.~P. Kingma and J.~Ba.
\newblock Adam: {A} method for stochastic optimization.
\newblock In Y.~Bengio and Y.~LeCun, editors, \emph{International Conference on
  Learning Representations}, 2015.
\newblock URL \url{http://arxiv.org/abs/1412.6980}.

\bibitem[Kipf and Welling(2017)]{kipf2017semi}
T.~N. Kipf and M.~Welling.
\newblock Semi-supervised classification with graph convolutional networks.
\newblock In \emph{International Conference on Learning Representations}.
  OpenReview.net, 2017.
\newblock URL \url{https://openreview.net/forum?id=SJU4ayYgl}.

\bibitem[Lee et~al.(2019)Lee, Sha, and Breazeal]{lee2019bayesian}
J.~J. Lee, F.~Sha, and C.~Breazeal.
\newblock A bayesian theory of mind approach to nonverbal communication.
\newblock In \emph{Proceedings of the ACM/IEEE International Conference on
  Human-Robot Interaction}, pages 487--496. IEEE, 2019.

\bibitem[Liu et~al.(2023)Liu, Zhu, Liu, Bisk, and Neubig]{liu2023computational}
A.~Liu, H.~Zhu, E.~Liu, Y.~Bisk, and G.~Neubig.
\newblock Computational language acquisition with theory of mind.
\newblock In \emph{International Conference on Learning Representations}, 2023.

\bibitem[Netanyahu et~al.(2021)Netanyahu, Shu, Katz, Barbu, and
  Tenenbaum]{netanyahu2021phase}
A.~Netanyahu, T.~Shu, B.~Katz, A.~Barbu, and J.~B. Tenenbaum.
\newblock Phase: Physically-grounded abstract social events for machine social
  perception.
\newblock In \emph{Proceedings of the aaai conference on artificial
  intelligence}, volume~35, pages 845--853, 2021.

\bibitem[Nguyen et~al.(2022)Nguyen, Nguyen, Le, Do, Venkatesh, and
  Tran]{nguyen2022learning}
D.~Nguyen, P.~Nguyen, H.~Le, K.~Do, S.~Venkatesh, and T.~Tran.
\newblock Learning theory of mind via dynamic traits attribution.
\newblock In \emph{Proceedings of the International Conference on Autonomous
  Agents and Multiagent Systems}, pages 954--962, 2022.

\bibitem[Nguyen et~al.(2023)Nguyen, Nguyen, Le, Do, Venkatesh, and
  Tran]{nguyen2023memory}
D.~Nguyen, P.~Nguyen, H.~Le, K.~Do, S.~Venkatesh, and T.~Tran.
\newblock Memory-augmented theory of mind network.
\newblock \emph{Proceedings of the AAAI Conference on Artificial Intelligence},
  37\penalty0 (10):\penalty0 11630--11637, Jun. 2023.
\newblock \doi{10.1609/aaai.v37i10.26374}.

\bibitem[Nguyen and Gonzalez(2020)]{nguyen2020cognitive}
T.~N. Nguyen and C.~Gonzalez.
\newblock Cognitive machine theory of mind.
\newblock In \emph{Proceedings of the Annual Meeting of the Cognitive Science
  Society}, 2020.

\bibitem[pandas~development team(2020)]{reback2020pandas}
T.~pandas~development team.
\newblock pandas-dev/pandas: Pandas, Feb. 2020.
\newblock URL \url{https://doi.org/10.5281/zenodo.3509134}.

\bibitem[Paszke et~al.(2019)Paszke, Gross, Massa, Lerer, Bradbury, Chanan,
  Killeen, Lin, Gimelshein, Antiga, et~al.]{paszke2019pytorch}
A.~Paszke, S.~Gross, F.~Massa, A.~Lerer, J.~Bradbury, G.~Chanan, T.~Killeen,
  Z.~Lin, N.~Gimelshein, L.~Antiga, et~al.
\newblock Pytorch: An imperative style, high-performance deep learning library.
\newblock \emph{Advances in Neural Information Processing Systems}, 32, 2019.

\bibitem[Pearson(1901)]{pearson1901liii}
K.~Pearson.
\newblock Liii. on lines and planes of closest fit to systems of points in
  space.
\newblock \emph{The London, Edinburgh, and Dublin philosophical magazine and
  journal of science}, 2\penalty0 (11):\penalty0 559--572, 1901.

\bibitem[Premack and Woodruff(1978)]{premack1978does}
D.~Premack and G.~Woodruff.
\newblock Does the chimpanzee have a theory of mind?
\newblock \emph{Behavioral and Brain Sciences}, 1\penalty0 (4):\penalty0
  515--526, 1978.

\bibitem[Puig et~al.(2020)Puig, Shu, Li, Wang, Liao, Tenenbaum, Fidler, and
  Torralba]{puig2020watch}
X.~Puig, T.~Shu, S.~Li, Z.~Wang, Y.-H. Liao, J.~B. Tenenbaum, S.~Fidler, and
  A.~Torralba.
\newblock Watch-and-help: A challenge for social perception and human-ai
  collaboration.
\newblock In \emph{International Conference on Learning Representations}, 2020.

\bibitem[Rabinowitz et~al.(2018)Rabinowitz, Perbet, Song, Zhang, Eslami, and
  Botvinick]{rabinowitz2018machine}
N.~Rabinowitz, F.~Perbet, F.~Song, C.~Zhang, S.~A. Eslami, and M.~Botvinick.
\newblock Machine theory of mind.
\newblock In \emph{International Conference on Machine Learning}, pages
  4218--4227. PMLR, 2018.

\bibitem[Sclar et~al.(2022)Sclar, Neubig, and Bisk]{sclar2022symmetric}
M.~Sclar, G.~Neubig, and Y.~Bisk.
\newblock Symmetric machine theory of mind.
\newblock In \emph{International Conference on Machine Learning}, pages
  19450--19466. PMLR, 2022.

\bibitem[Shu et~al.(2021)Shu, Bhandwaldar, Gan, Smith, Liu, Gutfreund, Spelke,
  Tenenbaum, and Ullman]{shu2021agent}
T.~Shu, A.~Bhandwaldar, C.~Gan, K.~Smith, S.~Liu, D.~Gutfreund, E.~Spelke,
  J.~Tenenbaum, and T.~Ullman.
\newblock Agent: A benchmark for core psychological reasoning.
\newblock In \emph{International Conference on Machine Learning}, pages
  9614--9625. PMLR, 2021.

\bibitem[Srivastava et~al.(2014)Srivastava, Hinton, Krizhevsky, Sutskever, and
  Salakhutdinov]{srivastava2014dropout}
N.~Srivastava, G.~Hinton, A.~Krizhevsky, I.~Sutskever, and R.~Salakhutdinov.
\newblock Dropout: a simple way to prevent neural networks from overfitting.
\newblock \emph{The Journal of Machine Learning Research}, 15\penalty0
  (1):\penalty0 1929--1958, 2014.

\bibitem[Takmaz et~al.(2023)Takmaz, Brandizzi, Giulianelli, Pezzelle, and
  Fernandez]{takmaz2023speaking}
E.~Takmaz, N.~Brandizzi, M.~Giulianelli, S.~Pezzelle, and R.~Fernandez.
\newblock Speaking the language of your listener: Audience-aware adaptation via
  plug-and-play theory of mind.
\newblock In \emph{Findings of the Association for Computational Linguistics:
  ACL 2023}, pages 4198--4217, Toronto, Canada, July 2023. Association for
  Computational Linguistics.
\newblock URL \url{https://aclanthology.org/2023.findings-acl.258}.

\bibitem[Tomasello(2010)]{tomasello2010origins}
M.~Tomasello.
\newblock \emph{Origins of human communication}.
\newblock MIT Press, 2010.

\bibitem[Ullman(2023)]{ullman2023large}
T.~Ullman.
\newblock Large language models fail on trivial alterations to theory-of-mind
  tasks.
\newblock \emph{arXiv preprint arXiv:2302.08399}, 2023.

\bibitem[Vaswani et~al.(2017)Vaswani, Shazeer, Parmar, Uszkoreit, Jones, Gomez,
  Kaiser, and Polosukhin]{vaswani2017attention}
A.~Vaswani, N.~Shazeer, N.~Parmar, J.~Uszkoreit, L.~Jones, A.~N. Gomez,
  {\L}.~Kaiser, and I.~Polosukhin.
\newblock Attention is all you need.
\newblock \emph{Advances in Neural Information Processing Systems}, 30, 2017.

\bibitem[Virtanen et~al.(2020)Virtanen, Gommers, Oliphant, Haberland, Reddy,
  Cournapeau, Burovski, Peterson, Weckesser, Bright, {van der Walt}, Brett,
  Wilson, Millman, Mayorov, Nelson, Jones, Kern, Larson, Carey, Polat, Feng,
  Moore, {VanderPlas}, Laxalde, Perktold, Cimrman, Henriksen, Quintero, Harris,
  Archibald, Ribeiro, Pedregosa, {van Mulbregt}, and {SciPy 1.0
  Contributors}]{virtanen2020scipy}
P.~Virtanen, R.~Gommers, T.~E. Oliphant, M.~Haberland, T.~Reddy, D.~Cournapeau,
  E.~Burovski, P.~Peterson, W.~Weckesser, J.~Bright, S.~J. {van der Walt},
  M.~Brett, J.~Wilson, K.~J. Millman, N.~Mayorov, A.~R.~J. Nelson, E.~Jones,
  R.~Kern, E.~Larson, C.~J. Carey, {\.I}.~Polat, Y.~Feng, E.~W. Moore,
  J.~{VanderPlas}, D.~Laxalde, J.~Perktold, R.~Cimrman, I.~Henriksen, E.~A.
  Quintero, C.~R. Harris, A.~M. Archibald, A.~H. Ribeiro, F.~Pedregosa, P.~{van
  Mulbregt}, and {SciPy 1.0 Contributors}.
\newblock {{SciPy} 1.0: Fundamental Algorithms for Scientific Computing in
  Python}.
\newblock \emph{Nature Methods}, 17:\penalty0 261--272, 2020.
\newblock \doi{10.1038/s41592-019-0686-2}.

\bibitem[Wang et~al.(2022)Wang, Zhong, Xu, and Wang]{wang2022tomc}
Y.~Wang, F.~Zhong, J.~Xu, and Y.~Wang.
\newblock {ToM2C}: {T}arget-oriented {M}ulti-agent {C}ommunication and
  {C}ooperation with {T}heory of {M}ind.
\newblock In \emph{International Conference on Learning Representations}, 2022.
\newblock URL \url{https://openreview.net/forum?id=2t7CkQXNpuq}.

\bibitem[{W}es {M}c{K}inney(2010)]{mckinney2010data}
{W}es {M}c{K}inney.
\newblock {D}ata {S}tructures for {S}tatistical {C}omputing in {P}ython.
\newblock In {S}t\'efan van~der {W}alt and {J}arrod {M}illman, editors,
  \emph{{P}roceedings of the 9th {P}ython in {S}cience {C}onference}, pages 56
  -- 61, 2010.
\newblock \doi{10.25080/Majora-92bf1922-00a}.

\bibitem[Wimmer and Perner(1983)]{wimmer1983beliefs}
H.~Wimmer and J.~Perner.
\newblock Beliefs about beliefs: Representation and constraining function of
  wrong beliefs in young children's understanding of deception.
\newblock \emph{Cognition}, 13\penalty0 (1):\penalty0 103--128, 1983.

\end{thebibliography}

\appendix

\section{Appendix} 

\subsection{CNN Architecture}
Our convolutional neural network (CNN) feature extractor consists of three convolutional layer blocks, followed by ReLU activation and max pooling.
The convolutional layers have 16, 32 and 64 filters, kernel size three and stride one.
For the max pooling layers we employ stride two.
The last max pooling layer is global.
This complete CNN stack is followed by a projection layer with size 64.

\subsection{\system for TBD}
The Triadic Belief Dynamics (TBD) dataset poses the task of predicting belief dynamics for five minds $m^i$, $i=\{1,2,12,21,c\}$, into four possible classes: \textit{appear}, \textit{disappear}, \textit{update} and \textit{null}.
In our \system models, we use a separate fully connected classification layer for each mind (cf.\ Figure 1 in the main paper).
Specifically, classification layers for $m^1$ and $m^{12}$ take $\bm{z}_1$ as input, whereas classification layers for $m^2$ and $m^{21}$ take $\bm{z}_2$ as input.
Since $m^c$ represents the common ground between both minds, $\bm{z}_1$ and $\bm{z}_2$ are first aggregated by performing element-wise multiplication and then fed into the classification layer.
Our Base-\system, by design, does not incorporate explicit ToM modelling. 
As a result, $\bm{z}_1$ and $\bm{z}_2$ are not computed, and the predictions for $m^i$ are generated by feeding the fully connected layers with $\bm{h}_{1}$ and $\bm{h}_{2}$. 

\subsection{Model Parameter Counts}

\begin{table}[h]
    \centering
    \begin{tabular}{l rr}
    \toprule
    \multirow{2}{*}{\textbf{Model}} & \multicolumn{2}{r}{\textbf{Number of parameters}} \\
    \cmidrule{2-3}
    & \textbf{BOSS} & \textbf{TBD} \\
    \midrule
    CNN & 21,411,270 & -- \\
    CNN+LSTM & 24,166,550 & -- \\
    CNN+GRU & 23,473,302 & -- \\
    CNN+Conv1D & 23,544,470 & -- \\
    \midrule
    CNN & -- & 162,457 \\
    CNN+HOG-LSTM & -- & 1,254,997 \\
    CNN+HOG+Mem & -- & 164,057 \\
    Feats+Mem & -- & 985,024 \\
    \midrule
    -MToMnet & & \\
    \addlinespace
    Base & 452,374 & 465,716 \\ 
    DB & 452,374 & 465,716 \\
    CG$\parallel$ & 460,886 & 474,228 \\
    CG$\oplus$ & 460,886 & 474,228 \\
    CG$\otimes$ & 460,886 & 474,228 \\
    CG$\odot$ & 493,654 & 506,996 \\
    IC$\parallel$ & 452,630 & 465,972 \\
    IC$\oplus$ & 452,630 & 465,972 \\
    IC$\otimes$ & 452,630 & 465,972 \\
    IC$\odot$ & 485,398 & 498,740 \\
    \bottomrule
    \end{tabular}
    \caption{Number of Parameters for models evaluated on BOSS and TBD.}
    \label{tab:params}
\end{table}

Table~\ref{tab:params} shows the number of parameter for each deep learning baseline we compare to and for each of our \system variant. 
As can be seen from the table, our \system has a significantly smaller number of parameters when compared to the baselines. 
This discrepancy becomes especially noticeable in the case of BOSS, as our \system variants have less than $3$\% of the parameters compared to the baselines (CNN, CNN+LSTM, CNN+GRU, CNN+Conv1D) -- approximately $460$k versus $21$M.

Baselines for TBD are based on a ResNet50, which has approximately $23$ million parameters. 
However, most of the ResNet50 layers are frozen, reducing the count of trainable parameters. 
Our \system has approximately one third of the parameters in CNN+HOG-LSTM and one half of Feats+Mem's parameters. 
CNN and CNN+HOG+Mem have less trainable parameters than \system, but they also achieve much lower F1 scores (see Table 1, main paper).

\subsection{Code}
Our code and bounding box annotations are available at \url{https://git.hcics.simtech.uni-stuttgart.de/public-projects/mtomnet}.

\subsection{Infrastructure \& Tools}
We ran our experiments on a server running Ubuntu 22.04, equipped with NVIDIA Tesla V100-SXM2 GPUs with 32GB of memory and Intel Xeon Platinum 8260 CPUs. 
We trained our models using PyTorch~\citep{paszke2019pytorch} with \texttt{1}, \texttt{42} and \texttt{123} as random seeds. 
All models were trained on a single GPU card, using a batch size of four for BOSS and $64$ for TBD. 
We used the Adam optimiser~\citep{kingma2015adam} with $\beta_1 = 0.9$, $\beta_2 = 0.99$, $\epsilon = 10^{-8}$, and a learning rate of $\eta = 5\cdot10^{-4}$. 
We did not perform any exhaustive hyper-parameter tuning as we noticed small variability for different configurations. 

We performed our data analysis using NumPy~\citep{harris2020array}, Pandas~\citep{reback2020pandas, mckinney2010data}, and SciPy~\citep{virtanen2020scipy}. 
Figures were made using Matplotlib~\citep{Hunter2007matplotlib}.

\subsection{Modelling Person-Specific Features -- Additional PCA Examples} 
\begin{figure*}
    \centering
    \includegraphics[width=\textwidth]{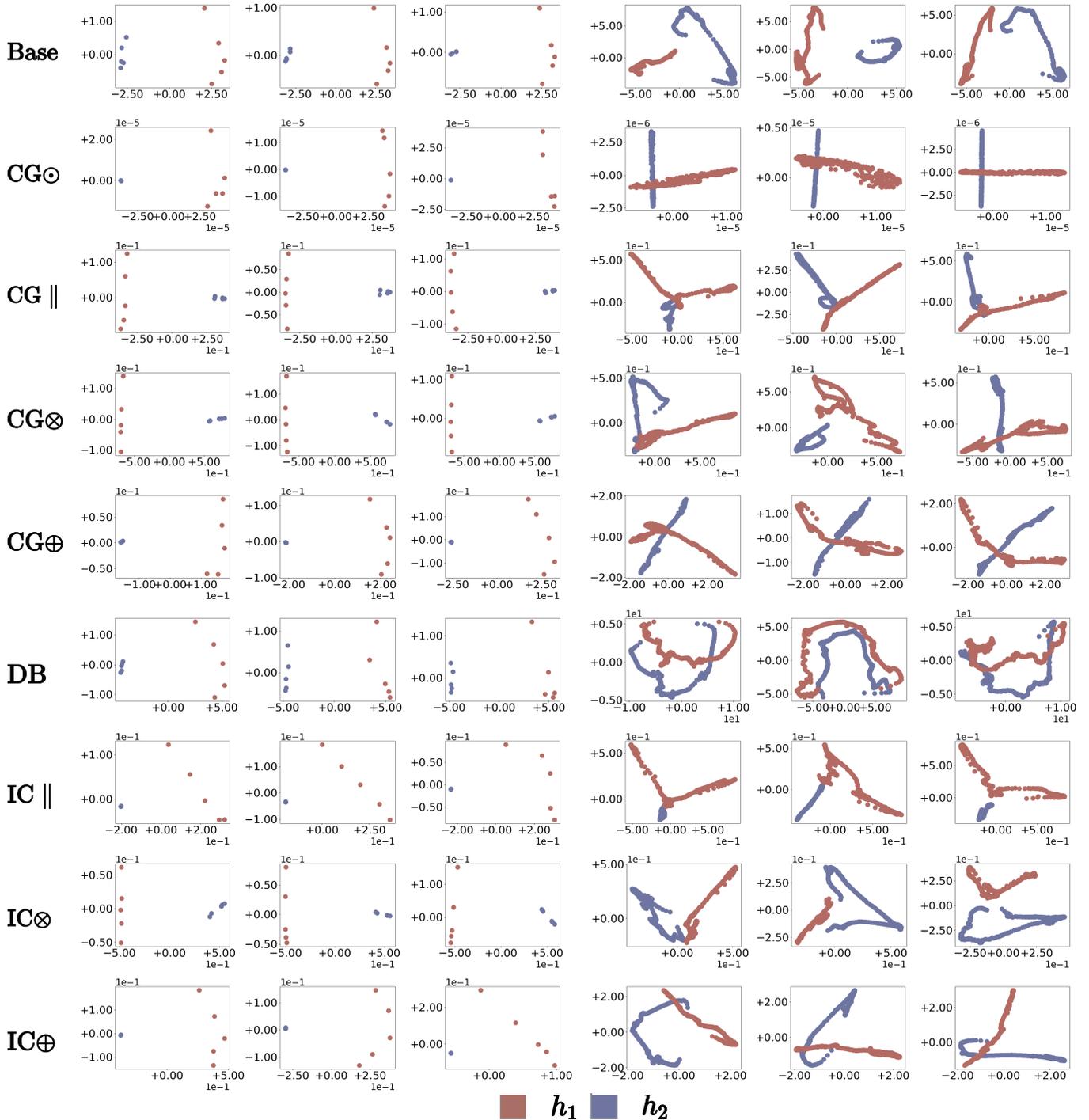}
    \caption{Examples from PCA results for our \system variants' latent states $\bm{h_1}$ and $\bm{h_2}$, taken from both TBD~\citep{fan2021learning} and BOSS~\citep{duan2022boss} datasets.}
    \label{fig:more_pca}
\end{figure*}

Figure~\ref{fig:more_pca} shows additional examples of PCA on the latent representations $\bm{h}_1$ and $\bm{h}_2$ for each of our \system variants. As the plots show, there is a clear disentanglement of principal components both on BOSS and TBD, for all \system variants.

\subsection{Ethical Impact}
While our work lays the foundation and remains distant from specific applications or immediate societal impact, we acknowledge that assertions related to modelling and predicting mental states can carry substantial ethical implications.
Mishandling such information has the potential to result in the inappropriate use of personal data,  reinforce biases or misunderstand complex psychological traits, potentially leading to unintended consequences on individuals' well-being.

\end{document}